\documentclass[a4paper,fleqn]{cas-sc}

\usepackage[round,authoryear]{natbib}
\usepackage{subfigure}
\usepackage{amsmath}
\usepackage{soul}
\usepackage{algorithm}
\usepackage{algorithmic}
\usepackage{bm}
\usepackage{color, xcolor}
\soulregister\cite7
\soulregister\citep7
\soulregister\ref7

\def\tsc#1{\csdef{#1}{\textsc{\lowercase{#1}}\xspace}}
\tsc{WGM}
\tsc{QE}

\begin{document}
\let\WriteBookmarks\relax
\def\floatpagepagefraction{1}
\def\textpagefraction{.001}

\shorttitle{Z. Yang et~al./ Expert Systems with Applications}
\shortauthors{Z. Yang et~al.}

\title [mode = title]{Neural Clothing Tryer: Customized Virtual Try-On via Semantic Enhancement and Controlling Diffusion Model}  

\author[1,2]{Zhijing Yang}
\ead{yzhij@gdut.edu.cn}

\author[1]{Weiwei Zhang}
\ead{2112303068@mail2.gdut.edu.cn}

\author[1]{Mingliang Yang}
\ead{ymlgdut@gmail.com}

\author[1]{Siyuan Peng}
\ead{peng0074@gdut.edu.cn}

\author[1]{Yukai Shi}
\ead{ykshi@gdut.edu.cn}

\author[3]{Junpeng Tan}
\ead{tjeepscut@gmail.com}

\author[1]{Tianshui Chen}[orcid=0000-0002-5848-5624]
\cormark[1]
\ead{tianshuichen@gmail.com}

\author[4]{Liruo Zhong}
\cormark[1]
\ead{Lee.Zhong@gsaitech.com}

\address[1]{Guangdong University of Technology, Guangzhou 510006, China}
\address[2]{Guangdong Provincial Key Laboratory of Intellectual Property \& Big Data, Guangzhou 510006, China}
\address[3]{South China University of Technology, Guangzhou 510006, China}
\address[4]{Genstoraige Technology (Beijing) Co., Ltd, Beijing 100000, China}
\cortext[1]{Tianshui Chen and Liruo Zhong are corresponding authors.}

\begin{abstract}
This work aims to address a novel Customized Virtual Try-ON (Cu-VTON) task, enabling the superimposition of a specified garment onto a model that can be customized in terms of appearance, posture, and additional attributes. Compared with traditional VTON task, it enables users to tailor digital avatars to their individual preferences, thereby enhancing the virtual fitting experience with greater flexibility and engagement. To address this task, we introduce a Neural Clothing Tryer (NCT) framework, which exploits the advanced diffusion models equipped with semantic enhancement and controlling modules to better preserve semantic characterization and textural details of the garment and meanwhile facilitating the flexible editing of the model's postures and appearances. Specifically, NCT introduces a semantic-enhanced module to take semantic descriptions of garments and utilizes a visual-language encoder to learn aligned features across modalities. The aligned features are served as condition input to the diffusion model to enhance the preservation of the garment's semantics. Then, a semantic controlling module is designed to take the garment image, tailored posture image, and semantic description as input to maintain garment details while simultaneously editing model postures, expressions, and various attributes. Extensive experiments on the open available benchmark demonstrate the superior performance of the proposed NCT framework.  
\end{abstract}





\begin{keywords}
  Customized Virtual Try-ON \sep Diffusion Model \sep Character Generation
\end{keywords}
\maketitle

\section{Introduction}
\begin{figure*}
  \centering
  \includegraphics[width=1\linewidth]{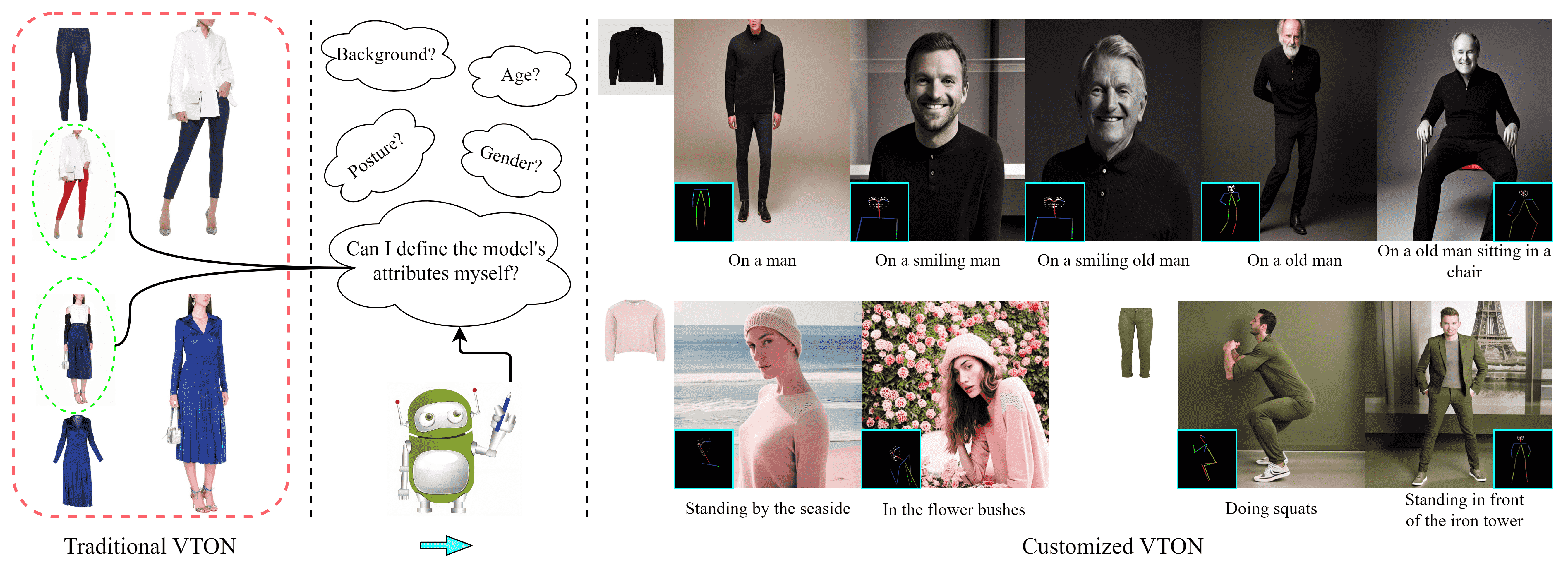}
  \caption{Illustration of the traditional and customized VTON tasks. Customized VTON devotes to superimpose a specified garment onto a model that can be customized in terms of appearance, posture, and additional attributes.}
  \label{fig:task}
\end{figure*}

Virtual Try-On (VTON) systems have significantly influenced the e-commerce fashion industry by offering an innovative solution for consumers to preview realistic try-on results of the target person wearing the given custom garment, while preserving the details of both humans and clothes \cite{LUO2024123213,mohammadi2021smart,VTON_TMM_2,VTON_TMM_3}. As shown in Fig. \ref{fig:task}, traditional VTON systems typically provide a standardized approach that lacks the capacity for deep customization, failing to account for the diverse facial appearances, postures, and other attributes. In pursuit of a more adaptive and user-centric solution, we introduce a novel Cu-VTON, aiming to transcend these limitations by allowing users to adjust digital avatars in terms of appearance, posture, and additional attributes \cite{lin2025geometry}, thereby providing a more flexible and engaging virtual Try-ON experience. 

As a powerful generative model, Generative Adversarial Networks (GANs) are widely used in various tasks \cite{TBD_1, TBD_2}. The core idea is to enable the generator to generate high-quality and realistic images through the adversarial training of the generator and the discriminator. The traditional VTON algorithms \cite{zhang2023limb,tsai2023attention,du2022vton,VTON_TMM_4} utilize the advantages of GAN and have made significant progress in the try-on task by introducing a generative adversarial network \cite{goodfellow2014generative,GAN_TMM_1} that is good at generating high-resolution images. However, these algorithms encounter difficulties in ensuring consistent representation of the garment across the diverse postures when applying to the Cu-VTON task. This arises from the inherent limitations associated with the mode collapse in adversarial training processes and the insufficient capability to capture and reproduce the intricate details and textures of clothing items. In recent years, a novel generative architecture known as diffusion models \cite{sohl2015deep,ho2020denoising,TMM_diffusion_1} has emerged, demonstrating exceptional quality in generation tasks coupled with a more stable training process. Inspired by their tremendous successes, existing works \cite{morelli2023ladi} introduce latent diffusion models extended with learnable skipping connections that can better preserve clothing details to address the traditional VTON task. Despite these advancements, such algorithms retain the intrinsic constraints of the VTON task, lacking the capability to adjust the models' postures and other attributes in a customized manner. Another stream of works \cite{ruiz2023dreambooth,hu2021lora} harnesses a collection of images to construct an embedding of a specific subject, which can be applied to generate novel images. These algorithms can adapt to address the Cu-VTON task by using the target clothing as the subject and superimposition of a specified garment onto a model whose appearance and posture can be edited. However, they lack the specialized mechanisms required to precisely manipulate clothing textures and details, and consequently tend to blur or distort the garment's features, leading to a compromise in the visual fidelity of the clothing item on the customized models.

In this work, we introduce an innovative NCT architecture that achieves the superimposition of a selected garment onto a model with customized features. This architecture harnesses the power of an advanced diffusion model \cite{ho2020denoising} for generating high-quality images and is augmented with semantic enhancement and controlling modules, enabling precise preservation of the semantic characterization and textural details of the garment and meanwhile facilitating the flexible editing of the model's appearance and posture. Instead of merely relying on visual clothing information, the semantic enhancement (SE) module further incurs semantic description for the target garment and exploits the visual-language encoder \cite{li2023blip} to learn aligned features for both the semantic description and visual information. These aligned features serve as a conditional input to the diffusion model to enhance the preservation of the garment’s semantics, and thus facilitate generating more reasonable fitting results across different postures. The semantic controlling (SC) module takes visual information and semantic description to control and edit the models from different perspectives. On one hand, a dual-branch controlling network takes the garment image and a specific posture as condition inputs to maintain garment details while simultaneously editing the model's posture. On the other hand, SC also takes semantic descriptions about expressions, ages, and other attributes as condition inputs for precise manipulation of corresponding attributes.

Our contributions in this work are threefold. Firstly, to our knowledge, this represents the inaugural effort to tackle the Customized Virtual Try-On task, which involves the precise superimposition of a chosen garment onto a model with customized options for appearance, posture, and other attributes. This approach promises a more dynamic and immersive virtual try-on experience. Secondly, we introduce a novel NCT architecture. Its key technical innovations include a new dual-branch conditioning mechanism that additively fuses decoupled garment and pose control signals, and a vision-centric semantic enhancement strategy that achieves a better interplay between detail preservation and attribute editing compared to existing methods. Lastly, we validate the efficacy of our proposed architecture through rigorous experiments on the widely-used Dress Code dataset \cite{morelli2022dress} and carry out a comprehensive set of ablative studies to quantify the individual contributions of each module.

\begin{figure*}
    \centering
    \includegraphics[width=1\linewidth]{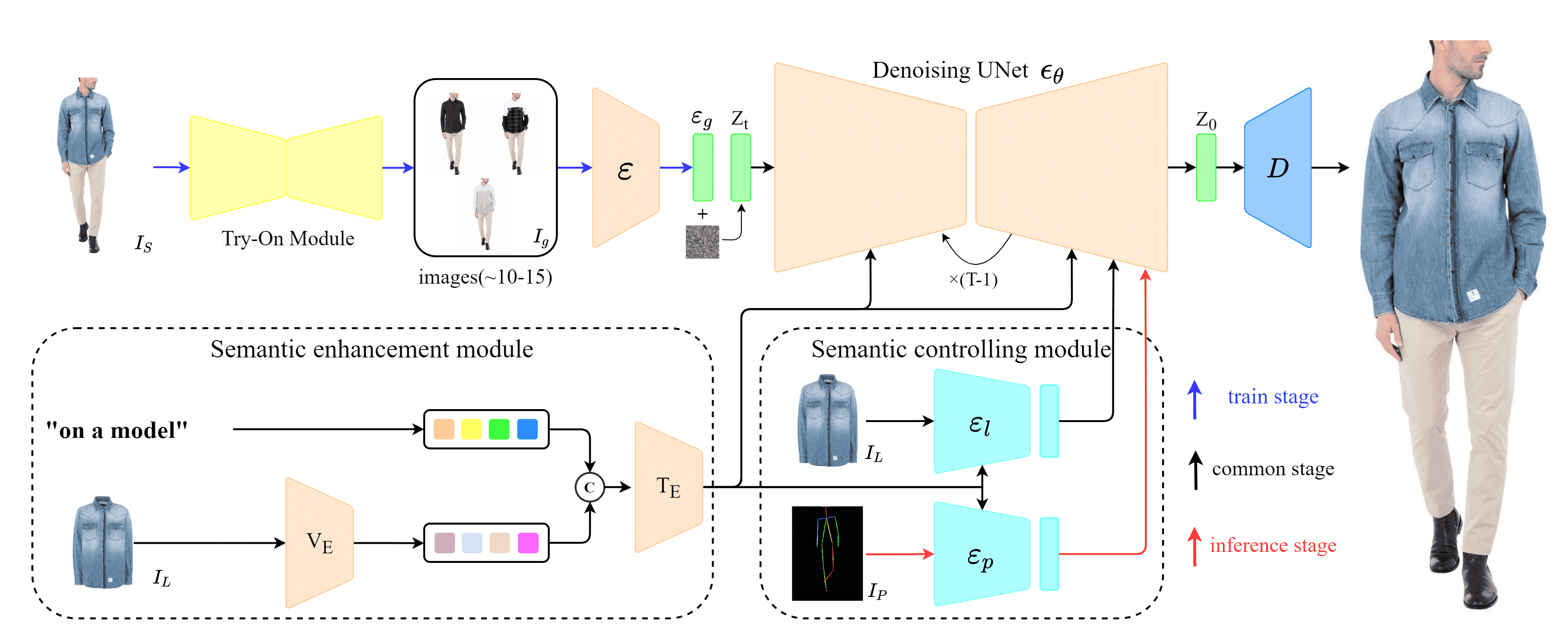}
    \caption{An overall illustration of the NCT framework. We equip the diffusion model with the SE and SC modules, in which SE learns aligned features of both semantic description and garment images as an enhanced condition to the diffusion models, while the SC takes garment images, posture information, and semantic description of other attributes as input to preserve the garment details and simultaneously edits posture and other attributes.}
    \label{fig:outline}
\end{figure*}

\section{Related Works}
\label{sec:Related}
In this section, we review the literature most relevant to our work, categorized into two primary domains that correspond to the baselines in our experiments: traditional image-based virtual try-on and subject-driven, text-conditioned image generation. For each category, we analyze its strengths and, more importantly, its limitations with respect to the proposed Customized VTON (Cu-VTON) task.

\subsection{Traditional Image-based Virtual Try-On}
In the past decade, image-based virtual try-on (VTON) techniques have been developed rapidly to enhance the online shopping experience by generating a fitting image of a person wearing a given garment \cite{li2023virtual, Virtual_Try-On_liusi}. Most of the foundational and subsequent methods were based on Generative Adversarial Networks (GANs) \cite{goodfellow2014generative,karras2020analyzing}. Seminal works include the original VTON framework \cite{han2018viton}, which utilized a coarse-to-fine pipeline, and ClothFlow \cite{han2019clothflow}, which adopted an appearance-flow-based model for better garment warping. Following these, a multitude of improved versions have been widely proposed, such as PL-VTON \cite{zhang2023limb}, POVNet \cite{li2023povnet}, HR-VTON \cite{lee2022high}, Clothformer \cite{jiang2022clothformer}, TryOnGAN \cite{lewis2021tryongan}, context-driven VTON \cite{fele2022c}, general-purpose VTON \cite{xie2023gp}, and size-aware VTON \cite{chen2023size}. However, due to the inherent weaknesses of the GAN model, these methods often suffer from drawbacks such as unrealistic results and poor detail preservation. More recently, to enhance generation quality and stability, the field has shifted towards diffusion models. For example, Morelli et al. \cite{morelli2023ladi} proposed LaDI-VTON, which incorporated a novel autoencoder into a latent diffusion model to better maintain garment details. Gou et al. \cite{gou2023taming} proposed a diffusion-based conditional inpainting method, while Zhu et al. \cite{zhu2023tryondiffusion} developed TryOnDiffusion, which adopted two UNets to preserve garment details during significant pose and body changes.

Despite their impressive advancements in generating high-fidelity try-on results, all these traditional VTON methods share a fundamental limitation: they operate on a fixed input person image. Consequently, while they excel at garment detail preservation for a static character, they inherently lack the capability for pose editing and attribute control. The model's identity, posture, and expression are predetermined by the input image and cannot be customized by the user. This rigidity prevents them from fulfilling the core requirements of our dynamic and flexible Cu-VTON task.

\subsection{Subject-Driven and Text-Conditioned Image Generation}
Another highly relevant line of work involves generating customized images based on textual descriptions \cite{xiaodan_Body_Generation,T2I_1,T2I_2,T2I_3,T2I_4,T2I_5,T2I_6,T2I_7, chen2024dynamic, chen2024heterogeneous}. In the era of diffusion models, the subfield of subject-driven (or theme-driven) generation has gained significant traction. These methods aim to learn a novel concept from a few images and synthesize it in new contexts. Prominent approaches like Textual Inversion \cite{Textual_inversion} use placeholders to represent visual concepts, while DreamBooth \cite{ruiz2023dreambooth} fine-tunes the diffusion model itself for better thematic fidelity. The introduction of parameter-efficient fine-tuning techniques like LoRA \cite{lora} has significantly reduced the cost of this process. These methods can be adapted for our task by treating the target garment as the "subject" to be learned and subsequently generated in new scenes, as demonstrated by other thematic works like character dance video generation \cite{maorui_Body_Generation}.

The primary strength of these methods lies in their powerful customization capabilities; they offer excellent pose editing and attribute control, allowing the learned subject to be synthesized on various characters with different appearances, poses, and expressions as guided by text prompts \cite{chen2024learning, chen2025contrastive, xu2025exploiting}. However, they were not designed for the high-fidelity requirements of virtual try-on. Their principal weakness is a significant struggle with garment detail preservation. As these methods are optimized to capture the general "essence" of a theme, they often fail to reproduce the intricate characteristics and fine-grained details of a specific garment, sometimes even failing to recognize it as clothing. This loss of detail is a critical failure for a satisfactory try-on experience.

In summary, our review of the related work reveals a clear and compelling research gap. Traditional VTON methods ensure detail fidelity but lack customization, while subject-driven and text-conditioned generation methods offer customization but sacrifice detail fidelity. The proposed Neural Clothing Tryer (NCT) is the first framework designed to systematically bridge this gap. It introduces a unified architecture that is engineered to simultaneously achieve high-quality garment detail preservation, flexible pose editing, and versatile attribute control, thereby directly addressing the challenges of the novel Cu-VTON task.

\section{Neural Clothing Tryer}

NCT builds on an advanced diffusion model for generating high-quality images and is further equipped with the semantic enhancement and controlling modules to preserve the textural details of given garments and meanwhile facilitate editing the model's appearance and posture flexibly. Specifically, the Semantic Enhancement (SE) module exploits a large-scale visual-language model to learn aligned features of semantic description and garment images to better preserve garment semantics. the Semantic Controlling (SC) module takes garment image, posture image, and semantic description of other attributes as condition inputs to maintain garment details while simultaneously editing model postures, expressions, and various attributes. An overall pipeline is illustrated in Fig. \ref{fig:outline}.

\subsection{Semantic Enhancement Module}
To ensure garment details are faithfully preserved during the generation of customized characters, we propose the Semantic Enhancement (SE) module. In contrast to subject-driven methods that distill visual information into a new textual token, our SE module adopts a vision-centric approach to create a rich multimodal embedding that directly leverages deep visual features.

In the traditional VTON task, a popular approach involves deforming or warping the provided clothing items to match the individual's pose when trying on the clothing. The warped clothing is then overlaid on the person, effectively adhering to their body. However, this method is not applicable to the Cu-VTON task, which aims to generate personalized characters and facilitate adaptive try-on of given clothing to these personalized characters. In summary, the Cu-VTON task revolves around addressing two key challenges: reliable personalized character generation and maintaining clothing information consistency. We found that diffusion models are effective tools for generating customized characters. However, diffusion models struggle to seamlessly integrate clothing information into the diffusion process to achieve a realistic try-on experience. Sometimes, diffusion models may find it difficult to discern the information of the target clothing provided to us, and may even fail to recognize it as clothing. For example, if the given clothing has intricate textures or patterns, diffusion models may mistakenly perceive them as mere textures or patterns, thus missing the main clothing information. To address this issue, we propose using semantic descriptions of clothing to enhance clothing information.

Inspired by the Blip-Diffusion method, we utilize Blip2 to extract target clothing information that matches the semantic description. Considering the model's generalizability of semantic features across different target clothing, we adopt a consistent semantic description for all target clothing, namely "clothes". Specifically, given the target clothing $I_L$, Blip2 is first used as a visual-linguistic encoder $V_E$ to extract coarse-grained image features that are consistent with the semantic description, denoted as $S_L$. Then, the feature of the text prompt $P$, denoted as $S_t$, is combined with $S_L$ to obtain the combined features $\widehat{S}(P, S_t)$. Finally, through the CLIP text encoder $T_E$, the combined information encapsulating the target clothing features and text prompt features is obtained, serving as the guiding input for generating customized virtual try-on, denoted as $S^*$. The formula is represented as:
\begin{equation}S^{*}=T_{E}(\widehat{S}(p,I_{L}))\end{equation}

\subsection{Semantic Controlling Module}
To address the lack of editability in traditional VTON, our Semantic Controlling (SC) module introduces a novel dual-branch conditioning mechanism to enable flexible control over posture while simultaneously maintaining garment details. We utilize visual information and semantic descriptions to control and edit the model from different aspects. Firstly, we use visual information, such as the image of the target clothing and pose information image, for appearance-based editing of the customized character. Then, we use detailed semantic descriptions, such as the age, gender, and background of the customized character, to define additional attributes of the customized character.

\noindent\textbf{Appearance Control.} For the input visual information, a dual-branch control network is proposed to enhance clothing details while editing the model's posture. Specifically, considering the compatibility between the dual-branch control network and the diffusion model, a trainable ControlNet $ \varepsilon_l(\cdot, \theta) $ is combined with a pre-trained ControlNet $ \varepsilon_p(\cdot) $ as the dual-branch control network. First, $ \varepsilon_l(\cdot, \theta) $ is introduced to extract fine-grained features of the target clothing image $ I_L $, calculated as:
\begin{equation}
y_l = \varepsilon_l(I_L, S^*, \theta)
\end{equation}

where $ y_l $ is the residual of clothing information condition added by the residual in the diffusion model's sampling and up-sampling blocks. Then we use the pre-trained pose control network $ \varepsilon_p(\cdot) $ to obtain the pose information from the pose image $ I_P $, calculated as:
\begin{equation}
y_p = \varepsilon_p(I_P, S^*)
\end{equation}

where $ y_p $ is the residual of posture information condition added by the residual in the diffusion model's sampling and up-sampling blocks. Finally, we combine $ y_l $ and $ y_p $, and jointly guide the clothing details and human pose in the diffusion model, enabling the model to effectively inherit the generative capabilities of the original diffusion model. Thus, the output of the dual-branch control network is:
\begin{equation}
y^*= y_l+y_p
\end{equation}

\noindent\textbf{Semantic Control.} Through the interaction between the defined "prompt" and the dual-branch control network, we can achieve control over other attributes by providing corresponding descriptive prompts along with skeleton images, for example, to control the age or expression of the customized character. With the powerful generation capability inherited from the original diffusion model, we can also use separate "prompts" to define other attributes, such as specifying the gender or background of the character.

\subsection{Training}
During the training of ${\varepsilon_l}(\cdot,\theta)$, unlike traditional VTON training, we do not need to perform clothing warping operations. We freeze all models except for the clothing detail control network and gradually add noise to the latent variable $\varepsilon_g$ of the person image $I_S$ to generate $\boldsymbol{z}_t$. Given a set of conditions, including the time step $t$ and the textual prompt $S^*$, the model learns a network $\epsilon_\theta$ to predict the noise added to $\boldsymbol{z}_t$. The corresponding objective function is defined as:
\begin{equation}\label{el_loss}
    L=\mathbb{E}_{\boldsymbol{z}_0, t, I_L, S^*, \epsilon \sim \mathcal{N}(0,1)}\left[\| \epsilon-\epsilon_\theta\left(z_t, t, I_L, S^*\right) \|_2^2\right]
\end{equation}
\begin{figure}
    \centering
    \includegraphics[width=0.5\linewidth]{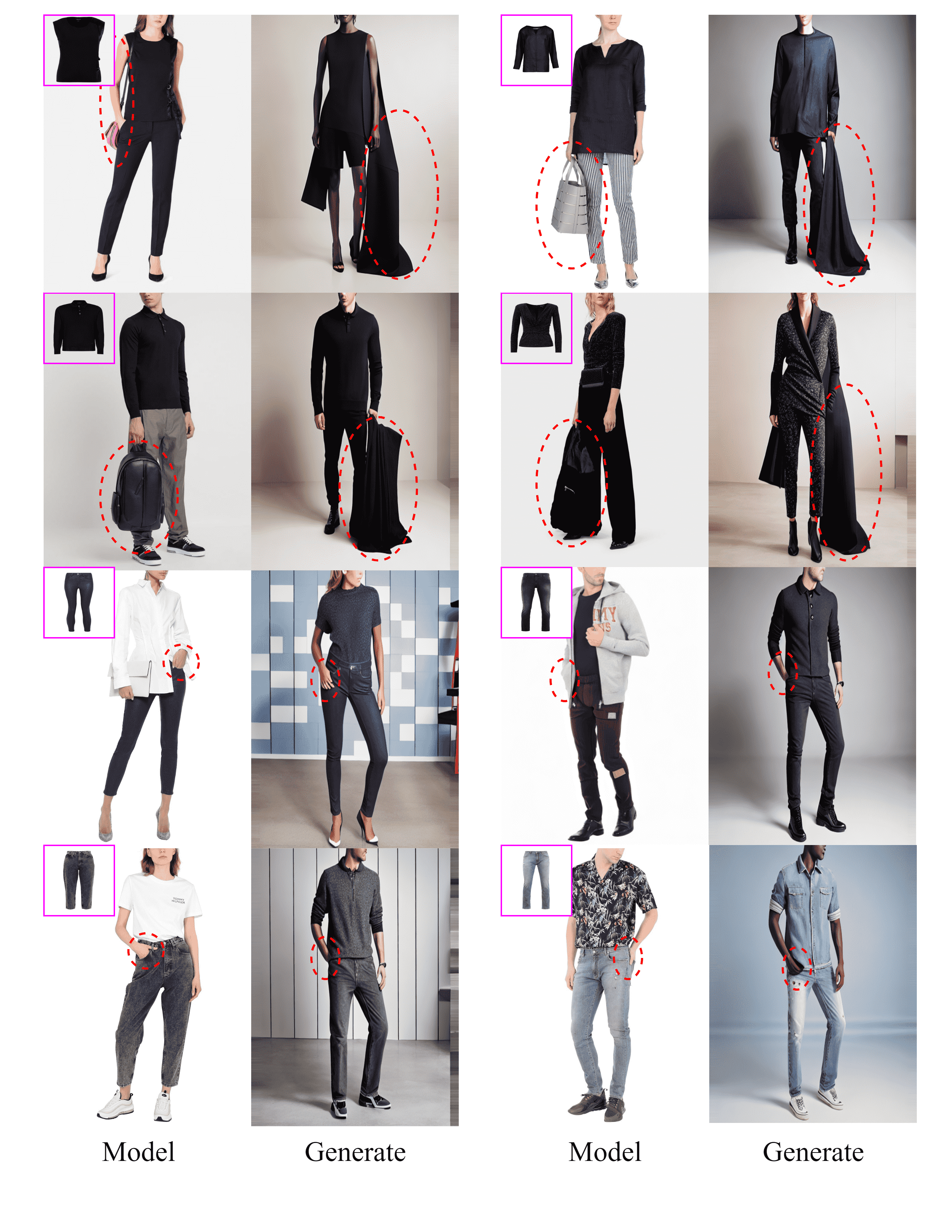}
    \caption{Comparison between model images in the Dress Code dataset and generation results obtained by NCT using the original Dress Code dataset for training.}
    \label{fig:IT_generate}
\end{figure}
However, a critical challenge we identified during initial training was the inherent "one-to-one pairing bias" within the Dress Code dataset. This bias refers to the phenomenon where a specific garment is almost exclusively paired with a single, unique model. When trained on such data, the network tends to learn spurious correlations between the garment's appearance and the idiosyncratic attributes of the model, such as their pose, body shape, hairstyle, or even their accessories like shoes and bags. This "memorization" behavior is highly detrimental, as it prevents the model from learning the generalizable, intrinsic features of the clothing. As illustrated in Fig. \ref{fig:IT_generate}, this bias leads to generated characters that unnaturally mimic the poses and accessories of the original training models, severely limiting the diversity and realism of the final output.

To overcome this limitation and foster true generalization, we introduce a data augmentation strategy based on cross-pairing synthesis. The core objective is to break the rigid one-to-one link and create a more diverse training distribution where the model can observe the same garment on different individuals. To implement this, we use a pre-existing traditional fitting model $\mathcal{F}$ as a "virtual tailor". The process involves first selecting a source person $I_S$ and multiple target garments ${\tilde{I}_L}^{1:k}$ from the dataset. Then, the model $\mathcal{F}$ fits each of the $k$ garments onto the single source person $I_S$, which synthesizes $k$ new person images ${I_g}^{1:k}$. This can be expressed by the formula:
\begin{equation}
{I_g}^{1:k} = \mathcal{F}(I_S, {\tilde{I}_L}^{1:k})
\end{equation}

By repeating this process across the dataset, we construct a new, augmented training set that effectively decouples garment features from personal attributes. Training our network on these diverse, synthetic images compels it to learn the person-agnostic essence of the clothing. This strategy effectively mitigates the bias brought about by one-to-one paired data training and promotes more diverse and realistic image generation results. The final objective function, trained on this augmented data, is given by:
\begin{equation}
    L=\mathbb{E}_{\boldsymbol{z}_0, t, I_L, S^*, \epsilon \sim \mathcal{N}(0,1)}\left[\| \epsilon-\epsilon_\theta\left(\tilde{\boldsymbol{z}}_t, t, I_L, S^*\right) \|_2^2\right]
\end{equation}

\section{Experiments}

\subsection{Experiment Settings}

\noindent\textbf{Implementation Details.} In our experiments, the training of the clothing details control network was conducted using the preprocessed Dress Code dataset, employing the Adam optimizer with a batch size of 2 for a total of 60,000 iterations. The learning rate was configured to $10^{-5}$, and a weight decay of $10^{-8}$ was applied.
To expedite the training process and efficiently manage memory utilization, we implemented mixed precision techniques \cite{mixed_precision, TBD_3, TBD_4} during both training and inference phases. Additionally, for noise scheduling, we employed a 50-step DDIM (Data-Dependent Initialization for Models) approach \cite{DDIM} and set the guiding parameter ${\alpha}$ for the classifier-free component to a value of 7.5.

\noindent\textbf{Evaluation Metrics.}
For the Cu-VTON task, we utilize the following metrics to evaluate clothing fidelity, semantic relevance, and pose consistency. It is important to note our choice of metrics is tailored to the unique challenges of the Cu-VTON task. While pixel-wise metrics like LPIPS and SSIM are standard for evaluating image reconstruction fidelity, they fundamentally require a pixel-aligned ground truth (GT) image for comparison. Our Cu-VTON task, by definition, generates novel images with customized attributes (e.g., new poses, identities, expressions), for which no such GT exists. Therefore, applying these metrics would be inappropriate and yield misleading results. Instead, we adopt a suite of metrics capable of evaluating different facets of our generation task without requiring a GT reference. It is also important to note that the following CLIP-based metrics operate on different principles and numerical scales. 1) \textbf{CLIP-I score} \cite{ruiz2023dreambooth}: Quantifies garment fidelity by measuring the average pairwise cosine similarity between the CLIP embeddings of the generated images and the real images. A higher score indicates better fidelity in preserving the garment's visual identity. 2) \textbf{CLIP-T (Text Similarity) score} \cite{ruiz2023dreambooth}: Calculates the average cosine similarity between the image embeddings of the generated images and the text embeddings of the \textit{garment's description}. Its values are typically in the [-1, 1] range, where a higher score is better, indicating a stronger alignment with the garment's textual attributes. 3) \textbf{CLIP-S (CLIP Score)} \cite{CLIP_s}: Assesses the adherence between the generated image and the full textual prompt used for its generation (which includes person attributes). Its calculation is distinct from a direct cosine similarity and results in a different, larger numerical scale. For this metric, a higher score is also better, reflecting greater overall image-prompt consistency.  4) \textbf{Pose Distance (PD) metric} \cite{baldrati2023multimodal}: Quantifies the consistency between the given target pose and the generated image's pose by computing the distance between extracted human body keypoints. A lower value indicates higher accuracy in pose control. This metric is not applicable to traditional VTON baselines that do not support pose editing functionality.

\begin{figure}
    \centering
    \includegraphics[width=0.6\linewidth]{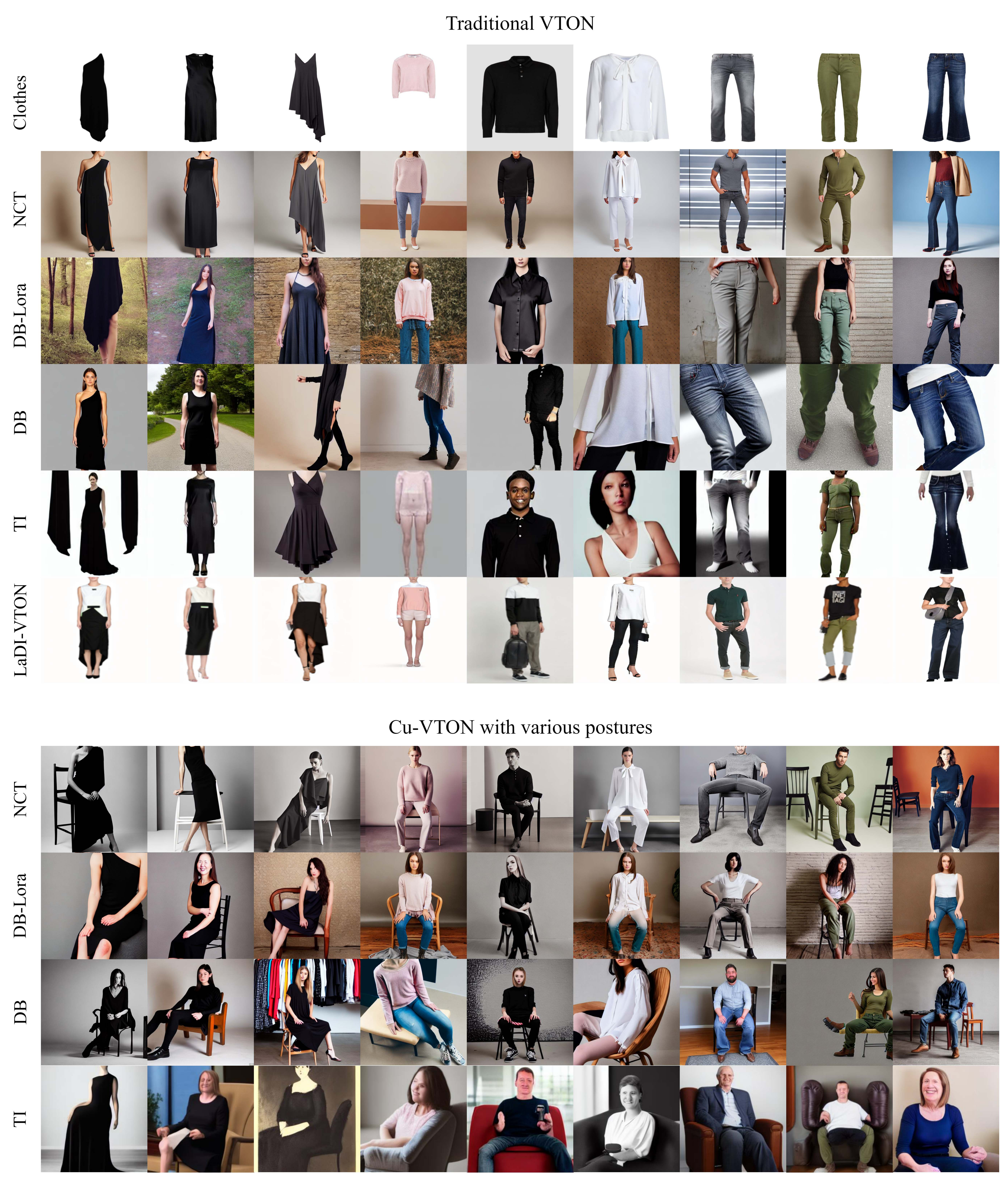}
    \caption{Visualization results of NCT and the competing methods on the Dress Code dataset. We present the results for both traditional VTON task with fixed posture and Cu-VTON task with various postures.}
    \label{fig:contrast}
\end{figure}

\begin{figure}
    \centering
    \includegraphics[width=0.6\linewidth]{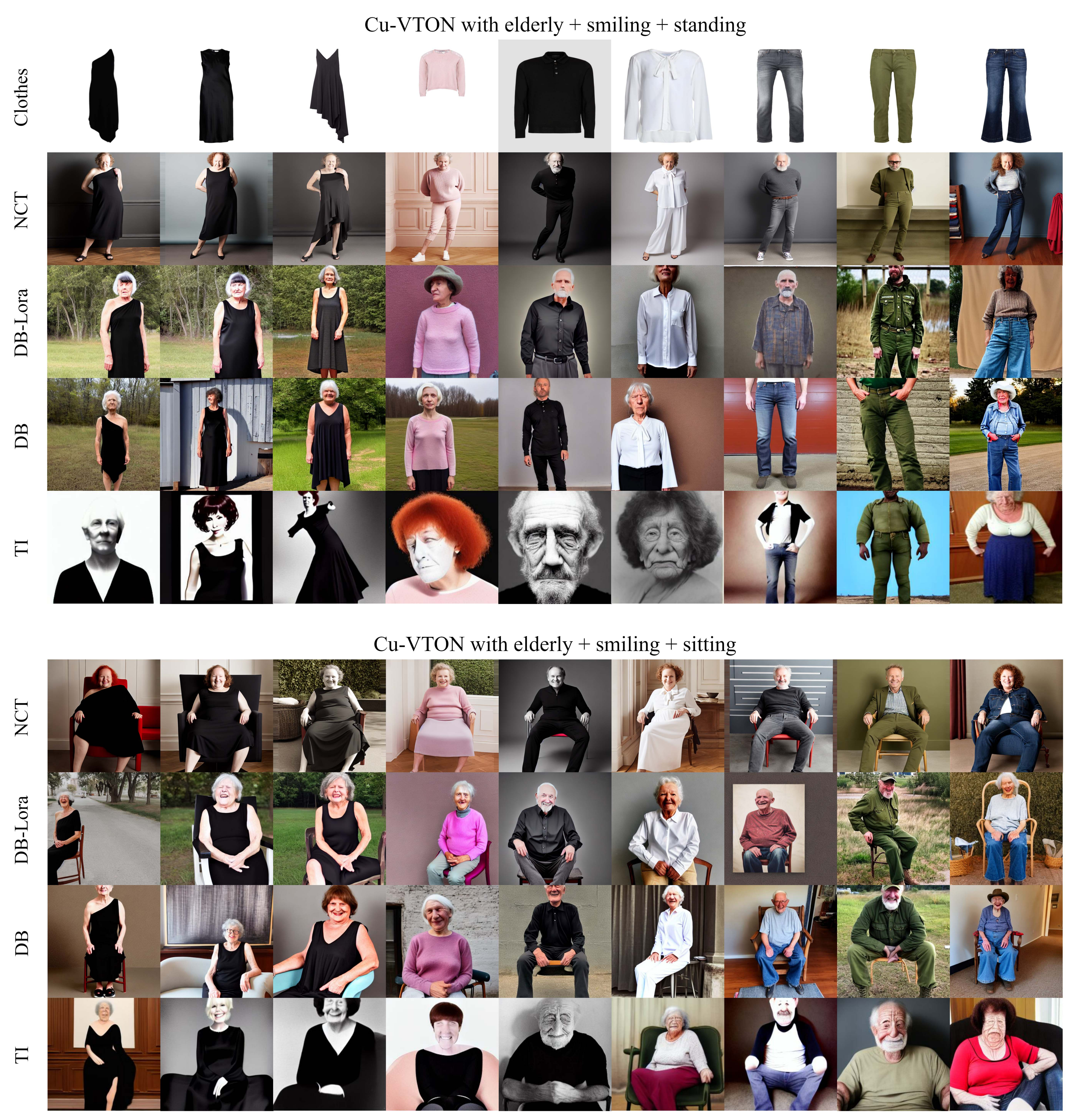}
    \caption{Visual comparison results of NCT and competitive methods on the Dress Code dataset. We present the generation results of NCT under various attribute editing conditions.}
    \label{fig:contrast3}
\end{figure}

\begin{figure}
    \centering
    \includegraphics[width=0.5\linewidth]{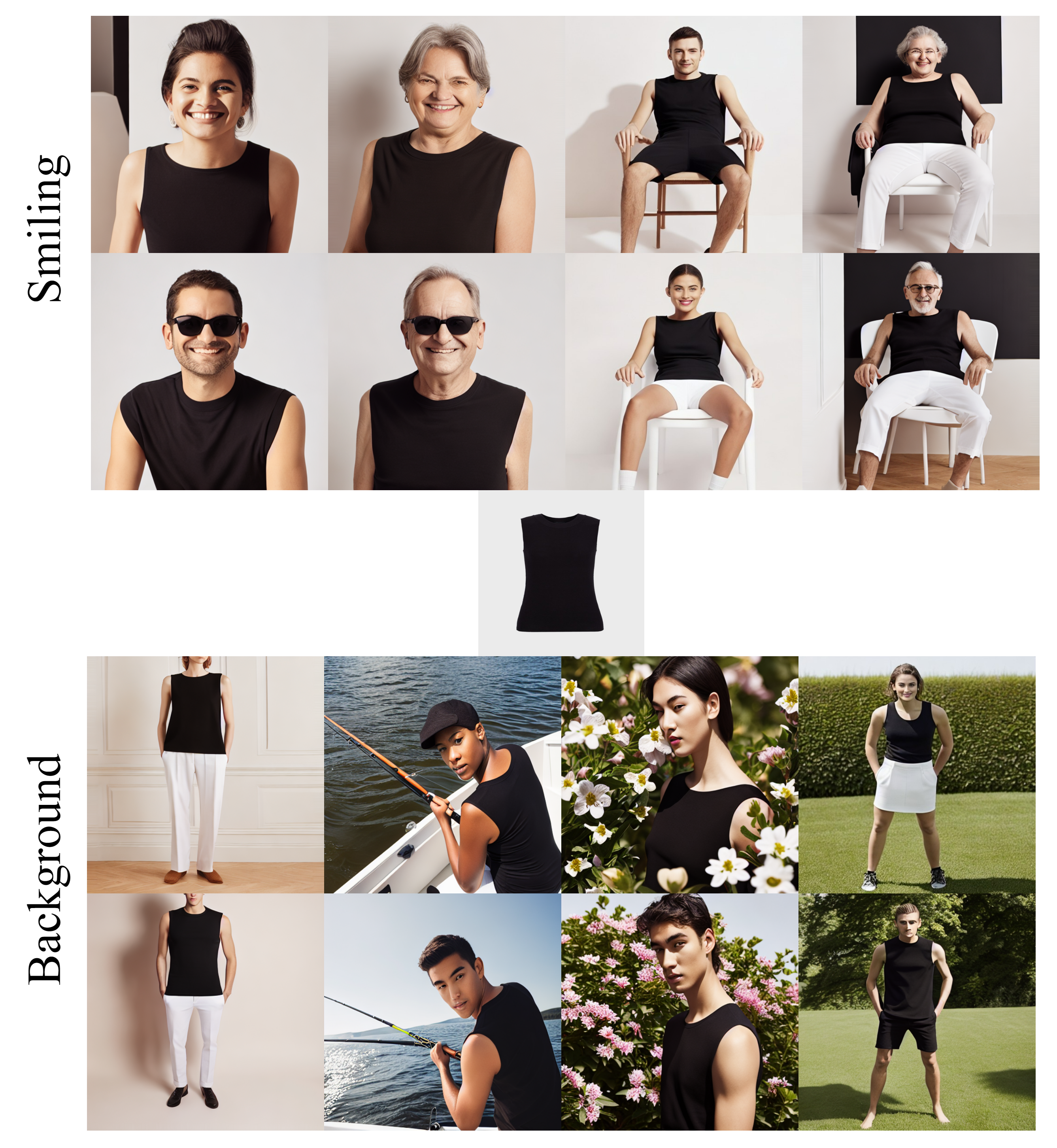}
    \caption{Various visual demonstrations generated by NCT using the same garment.}
    \label{fig:generate}
\end{figure}

\subsection{Comparison with Competitive Baselines}
Cu-VTON is an emerging task and thus no existing works to address this task are available for direct comparisons. In this work, we re-implement some competing baselines of two closely-related tasks: traditional VTON tasks, like LaDI-VTON \cite{morelli2023ladi} and theme-driven generation, like DreamBooth (DB) \cite{ruiz2023dreambooth}, DreamBooth augmented by low-rank optimization (DB-Lora) \cite{lora}, and Textual Inversion (TI) \cite{Textual_inversion}. We perform comparisons on both traditional VTON and cu-VTON settings. 

\noindent\textbf{Qualitative Comparison.} 
We present a visual comparison of the NCT method and competitive baselines in Fig. \ref{fig:contrast}. We conducted a qualitative comparison from two aspects. First, we evaluated the realism and naturalness of the generated clothing; the competitive baselines consistently failed to achieve high fidelity in the generated images. For instance, in the green trousers of the 8th column and the black top of the 5th column, the clothing generated by LaDI-VTON significantly mismatches the provided target clothing. DB and DB-Lora also struggled to accurately reproduce the given clothing, resulting in incorrect pants colors and inaccurate top buttons. Text inversion mostly failed to generate accurate images, and even the generated clothing lost the essence of the given clothing. Moreover, in traditional VTON experiments, DB and DB-Lora had difficulty dressing the customized character correctly in the target clothing. In the green trousers of the 8th column and the black top of the 5th column, the generated images did not fit the customized character well, leading to irregular deformations in the clothing. In contrast, the NCT method effectively integrates image features into the diffusion model, ensuring more realistic and natural fitting results. Secondly, we assessed the naturalness and completeness of the customized character; in traditional VTON tasks, the character is pre-set and cannot be customized. Therefore, this paper only discusses the results of the theme-driven customized character generation task. The NCT method can generate a customized character that fits seamlessly into the target clothing for fitting, while competitive baselines struggle to generate a customized character that exhibits natural and complete movement. For example, in the first and second columns involving clothing generation, DB-Lora failed to generate a complete and natural customized character, resulting in incomplete clothing being tried on. Similarly, in the seventh and eighth columns performing pants generation, DB and DB-Lora mainly only generated parts of the customized character wearing pants, neglecting other body parts.

Next, we compared the performance of NCT and other benchmark algorithms in multi-attribute editing. In Fig. \ref{fig:contrast3}, "Cu-VTON with elderly + smiling + standing" shows the visual results of NCT and competitive baselines generating characters with attributes of "elderly", "smiling", and "standing", while "Cu-VTON with elderly + smiling + sitting" shows the visual results of NCT and competitive baselines generating characters with attributes of "elderly", "smiling", and "sitting". First, we assessed the realism and naturalness of the generated clothing. The generation results of TI were poor in both realism and naturalness, with severe distortion when generating the upper body. DB and DB-Lora had a high degree of realism and naturalness in the generation of skirts and tops, but there were some issues, such as the incorrect button position generated by DB-Lora in the third row of the fifth column, and the same issue occurred in the seventh row of the same column. In the fourth and seventh rows of the first column, the skirt tail generated by DB was very poor, almost sticking the given target clothing onto the character, resulting in low naturalness of the clothing. DB and DB-Lora showed lower realism when generating pants, which may be related to the inherent memory of the diffusion model. When the given clothing does not match the attributes of the generated character, it leads to incorrect generation results. However, NCT overcame this limitation, ensuring high-quality fitting results. In terms of the naturalness and completeness of the customized character, the customized characters generated by competitive baselines were not as natural and realistic as those generated by NCT. This is due to NCT's effective use of semantic enhancement and semantic control, allowing the model to correctly recognize the relationship between the customized character and the clothing generation process. The smiles on the faces of the customized characters generated by NCT were also more natural.

Finally, in Fig. \ref{fig:generate}, we present a variety of visual demonstrations generated by NCT using the same piece of clothing. We used "smiling" as a prompt to generate customized characters of different genders, ages, and poses, and then created customized characters with different background prompts. The visual experimental results show that NCT can generate a diverse range of customized characters based on the given prompts while ensuring the consistency of the generated clothing with the provided clothing. This provides compelling evidence of NCT's outstanding ability in personalized generation.

\begin{table*}
\centering
\caption{CLIP-I, CLIP-T, CLIP-S, and PD comparisons of NCT and the competing methods on the Dress Code dataset.}
\resizebox{\linewidth}{!}{%
\begin{tabular}{ccccccccccccccc}
\hline
 ~     & \multicolumn{4}{c}{Dresses}    &~  & \multicolumn{4}{c}{Upper body}  &~  & \multicolumn{4}{c}{Lower body}                        \\ 
\cline{2-5} \cline{7-10} \cline{12-15}
 Model     & CLIP-I $\uparrow$     & CLIP-T $\uparrow$   &CLIP-S $\uparrow$  & PD$\downarrow$   &~  & CLIP-I $\uparrow$   & CLIP-T $\uparrow$  &CLIP-S $\uparrow$  & PD$\downarrow$   &~  & CLIP-I $\uparrow$   & CLIP-T $\uparrow$  &CLIP-S $\uparrow$  & PD$\downarrow$\\ 
\hline
NCT              & \textbf{0.766} & \textbf{0.254} & \textbf{19.672} & 1.64 &~ & \textbf{0.766} & \textbf{0.722} & \textbf{0.263}  & 0.174  &~ & \textbf{0.794} & \textbf{0.274} & \textbf{19.628}  & 2.92   \\
DB-Lora   & 0.734 & 0.249 & 17.064 & N/A &~ & 0.717 & 0.241 & 17.056  & N/A  &~ & 0.726 & 0.257 & 0.257  & N/A   \\
DB       &0.706 & 0.225 & 19.564 & N/A &~ & 0.705 & 0.186 & 16.989  & N/A  &~ & 0.747 & 0.216 & 17.147  & N/A   \\
TI & 0.643 & 0.247 & 16.915 & N/A &~ & 16.915 & 0.229 & 17.013  & N/A  &~ & 0.679 & 0.242 & 16.941  & N/A   \\
\hline
\end{tabular}
}
\label{Evaluation_Metrics}
\end{table*}

\noindent\textbf{Quantitative Comparison.} 
Table \ref{Evaluation_Metrics} provides an overview of the CLIP-S, CLIP-I, and CLIP-T scores obtained from our evaluation process. We thoughtfully selected ten distinct clothing items from each of the three categories featured in the Dress Code dataset, culminating in a total of 30 generated images for each item and yielding a dataset comprising 900 images in total. Subsequently, the evaluation metrics were computed based on this dataset, and the average scores are reported.
Overall, the results are consistent with visual observations. In the comparison of traditional VTON tasks, we focused on garment fidelity, specifically measuring the high or low scores of CLIP-I. NCT's superiority in CLIP-T scores indicates that the target garments generated by NCT are more similar to the given garments, demonstrating NCT's ability to ensure high-quality garment fidelity. In the comparison of theme-driven tasks, in addition to CLIP-I, we also considered CLIP-S and CLIP-T. We found that NCT consistently outperformed other comparative methods. This superiority emphasizes NCT's proficiency in preserving coarse-grained garment attributes and generating high-quality images. It also confirms that NCT provides impressive editability. Finally, for the Cu-VTON task, we provided PD scores for reference.

\begin{table*}
\centering
\caption{Clothing naturalness and Customizing accuracy of NCT and the competing methods by user studies. For all categories, the observed user preferences are statistically significant ($\chi^2$ test, p < 0.001).}
\resizebox{\linewidth}{!}{%
\begin{tabular}{cccccccccccc}
\hline
 ~     & \multicolumn{5}{c}{Clothing naturalness}    &~  & \multicolumn{5}{c}{Customizing accuracy}                       \\ 
\cline{2-6} \cline{8-12}
 Clothing categories     &NCT   & DB-Lora  &DB  & TI  & LaDI-VTON  &~  &NCT   & DB-Lora  &DB  & TI  & LaDI-VTON \\ 
\hline
Dresses    & 65.65\% & 21.9\%  & 3.15\%  &6.3\% & 3.15\% &~ & 72.93\%  & 16.7\%  & 6.27\%  &4.17\%  & N/A\\
Upper Body & 62.5\% & 17.27\% & 12.5\% & 6.27\%  & 1.5\% &~ & 62.53\%  & 15.7\%  &12.5\%  &4.2\%  & N/A \\
Lower Body &56.27\%  & 11.43\% & 18.8\% & 12.5\% & 1.1\% &~ & 68.8\%  & 16.7\%  & 18.8\%  &2.1\%  & N/A\\
\hline
\end{tabular}
}
\label{user_study}
\end{table*}

\noindent\textbf{User Study.}
We conducted a web-based user study to demonstrate the comparison between NCT and competitive baselines. It consists of two parts, corresponding to the aforementioned two metrics, namely garment fidelity and naturalness, as well as character customization naturalness and completeness. For each of the three Dress Code categories, we selected three garments. In the end, we obtained 45 images for evaluating garment fidelity and naturalness, and 36 images for evaluating character customization naturalness and completeness. We recruited 20 participants, and each participant was asked to make choices for these 81 images. The results, as shown in Table \ref{user_study}, indicate that NCT achieved the highest scores in both clothing ratings and customized character ratings. This demonstrates that NCT is capable of generating highly realistic clothing while producing reasonably natural customized characters.

\subsection{Ablative Studies}
\begin{figure}
    \centering
    \includegraphics[width=0.5\linewidth]{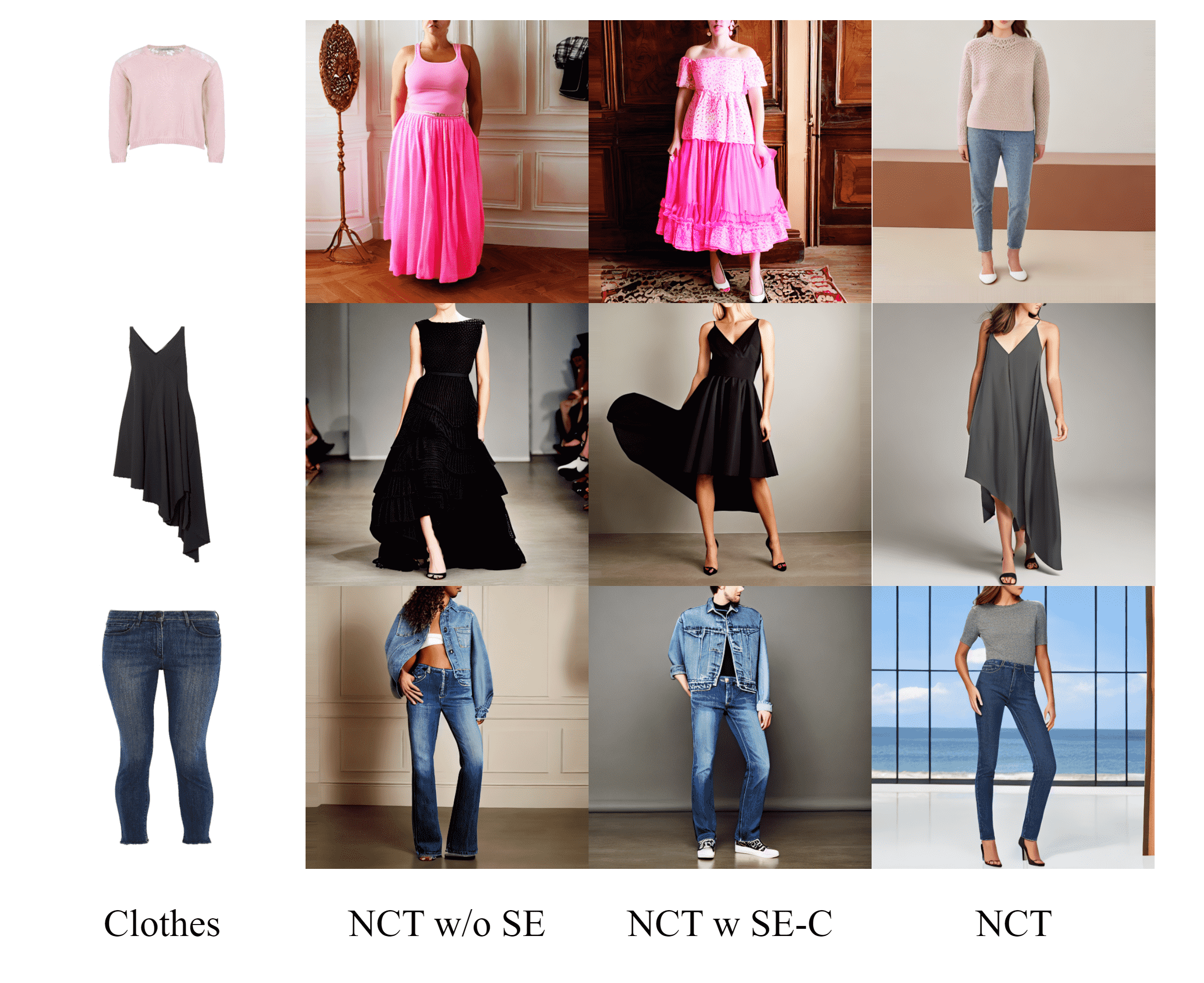}
    \caption{Comparisons of NCT, NCT without semantic enhancement (NCT w/o SE), and NCT with semantic enhancement by CLIP (NCT w/ SE-C).}
    \label{fig:Ablation-TSEM}
\end{figure}
\begin{figure}
    \centering
    \includegraphics[width=0.6\linewidth]{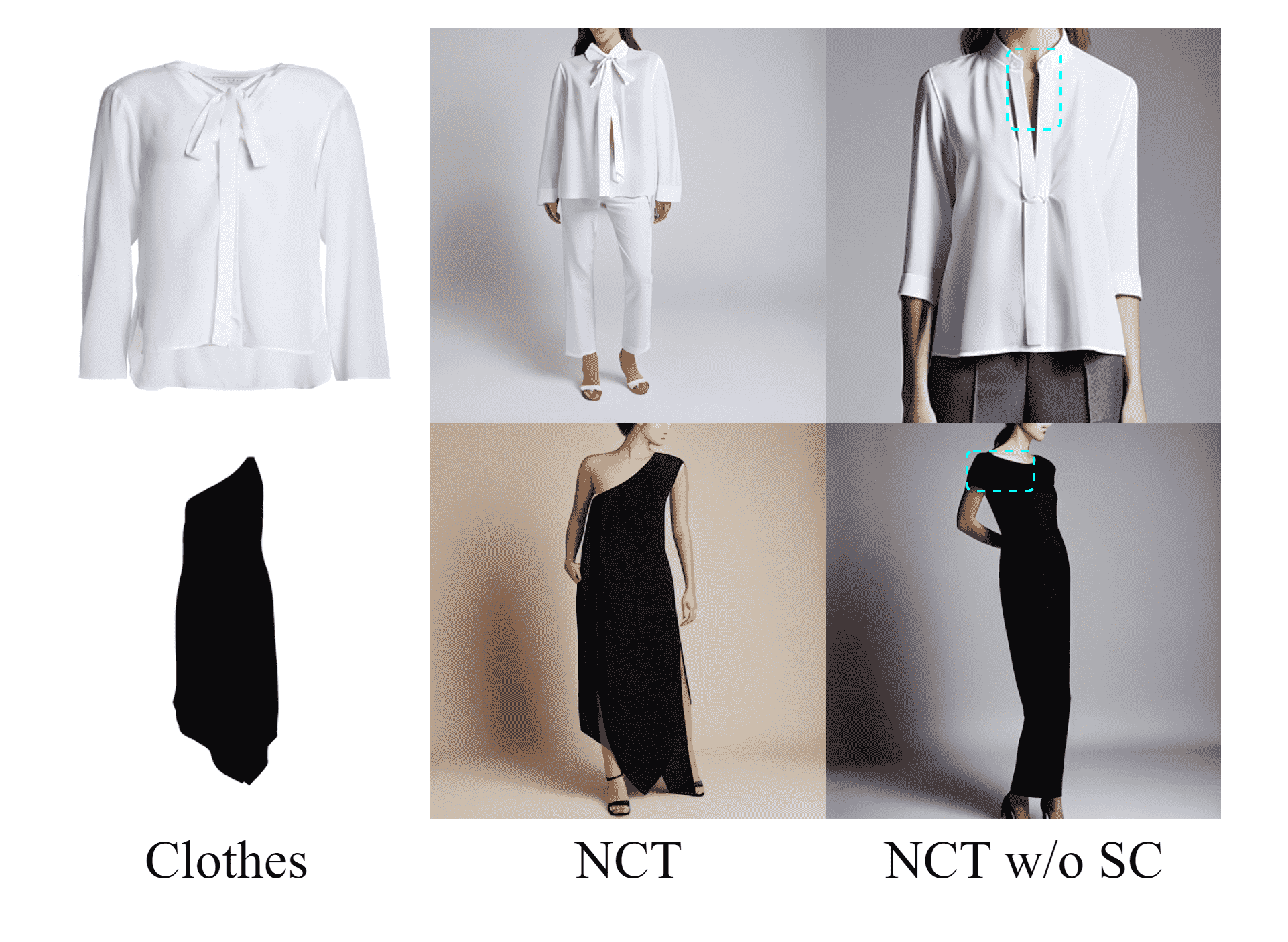}
    \caption{In the ablation experiment of SC, we compared the results between using and not using SC, and highlighted the errors in the generated clothing using dashed lines.}
    \label{fig:Ablation-DCM}
\end{figure}
\noindent\textbf{Analyses of Semantic Enhancement Module.}
In order to evaluate the impact of semantically aligned clothing features on preserving clothing information in NCT, we conducted an experiment where we removed SE and relied solely on text prompts to describe the clothing. This experiment is referred to as "NCT w/o SE". In Fig.  \ref{fig:Ablation-TSEM}, we observed significant differences between the generated clothing by the model and the provided clothing descriptions. Notably, in the first row, a long-sleeved top was incorrectly interpreted as a camisole, and the generated skirt exhibited noticeable deviations in style and color. Compared to the tops and skirts, the generated pants in the third row showed slightly better realism. This is because the pants had a common style, such as jeans, and the model's generation relied on the pre-existing diffusion models' capabilities rather than capturing specific features of the pants. This observation indicates that the lack of clothing information greatly limits the model's ability to faithfully reconstruct the target clothing. Relying solely on textual descriptions can only provide coarse-grained information, which has proven insufficient for capturing the intricate details required for accurate clothing synthesis.

Subsequently, we introduced the SE-C method, which utilizes the CLIP (Contrastive Language-Image Pretraining) approach instead of BLIP2 to extract features from clothing images, as a replacement for SE. Fig.  \ref{fig:Ablation-TSEM} provides a visual representation of the results obtained from these comparative experiments. We observed an improvement in the fidelity of the generated target clothing compared to the "NCT w/o SE" experiment. It successfully captured the basic features described in the provided clothing descriptions to some extent. For example, in the second row, when generating a skirt, the model correctly generates that the upper body of the skirt is a sleeveless style, and for pants, it correctly identified the leg portion as tight-fitting. However, challenges still exist in accurately reconstructing certain aspects, such as the long-sleeved attribute of the top and the style details of the short skirt hem. These imperfections can be attributed to using CLIP as the feature extractor, which introduces additional errors. CLIP is not explicitly tailored to the diffusion models, and thus, the clothing information extracted by CLIP may not align optimally with the diffusion models. This misalignment results in inaccurate clothing information being obtained by the NCT model, thereby reducing the fidelity of the generated clothing.
\begin{figure}
    \centering
    \includegraphics[width=0.6\linewidth]{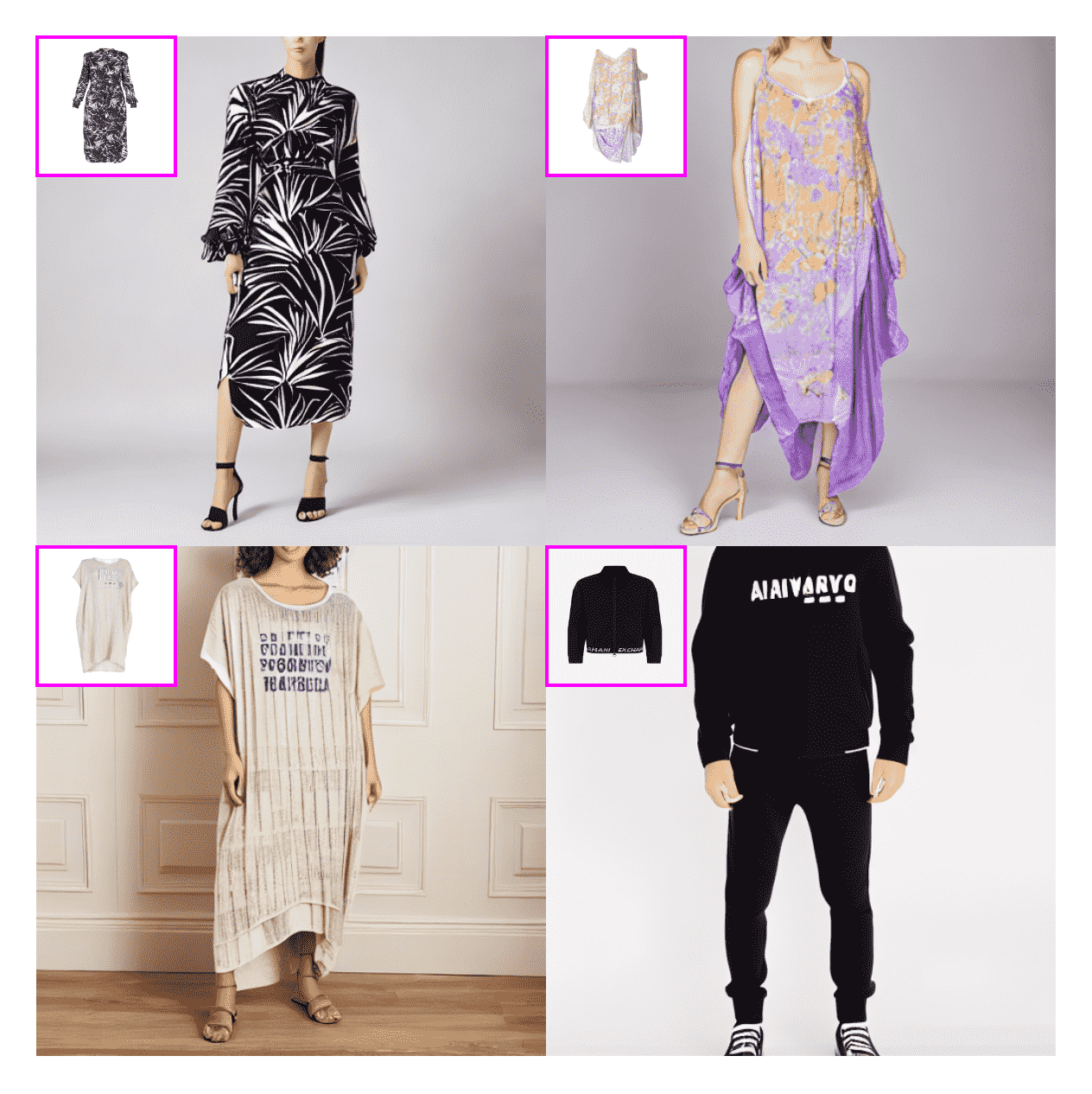}
    \caption{Several failure cases generated by the proposed NCT.}
    \label{fig:limitation}
\end{figure}

\noindent\textbf{Analyses of Semantic Controlling Module.}
In order to evaluate the role of SC for supplementary garment generation details and pose control in NCT, we intentionally conducted experiments in which this module was excluded. Fig.  \ref{fig:Ablation-DCM} provides a visual representation of the experimental results, clarifying the significant impact of missing SC on the retention of fine-grained features in the generated clothing images. In the generated white blouse image, "NCT w/o SC" does not generate the collar in the original blouse, and the single off-shoulder dress is also generated as a traditional shoulder-length dress. This indicates that the generated images fail to accurately capture and preserve the complex details of the target clothing.

In addition, missing SC will lead to randomness in the generated poses. Relying on descriptions alone to generate the expected pose well introduces irregular deformations to the human subject. More importantly, costume generation and human pose adjustment should be complementary, and in the case of "NCT w/o SC", the generated skirt tends to conform to the body pose, resulting in the skirt looking tightly wrapped around the character. This random pose generation weakens the naturalness of the generated clothing.

\begin{figure}
    \centering
    \includegraphics[width=0.6\linewidth]{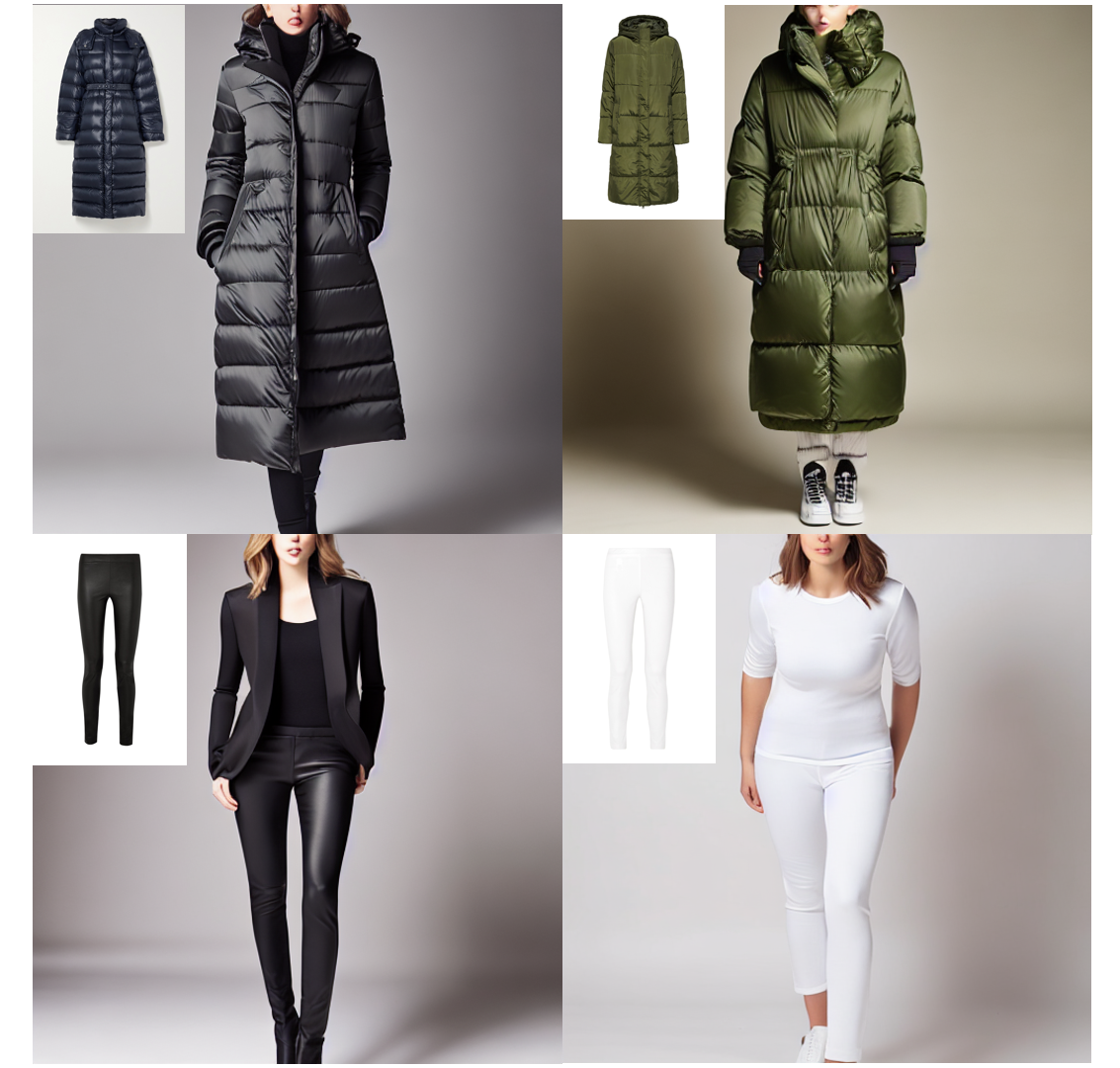}
    \caption{Qualitative results of NCT on special, out-of-distribution clothing types.}
    \label{fig:heavy_coat_tights}
\end{figure}

\subsection{Limitation}
Although the NCT has produced high-quality customized virtual fitting results, there are still inevitable limitations.
First, as can be seen from Fig. \ref{fig:contrast} and \ref{fig:contrast3}, the quality of the fingers generated by the NCT when creating custom characters is sometimes low. This is a well-known issue stemming from the inherent limitations of many diffusion models. While this could be resolved by simply replacing the backbone with a more effective diffusion model, a more promising future direction involves a two-stage refinement strategy, where a specialized, lightweight model is used to refine critical areas like hands and faces after the initial generation. Second, the model's ability to generalize to garments with highly complex details or out-of-distribution characteristics is currently limited. As shown in Fig. \ref{fig:limitation}, the NCT already finds it challenging to perfectly generate complex patterns and text on clothing. This is because relying solely on the semantic description from our SE module, while robust, can lead to the loss of such high-precision information. To further investigate this boundary, we conducted preliminary tests on special clothing types not well-represented in the training data (see Fig. \ref{fig:heavy_coat_tights} for examples). The results show mixed performance. For garments with special materials like heavy coats or tights, the model can often preserve the basic silhouette and capture some macro-level properties (e.g., heavy folds, a subtle sheen). However, it struggles with geometric consistency in local areas, such as the collar of the coat, producing noticeable artifacts. For garments with extremely intricate patterns, such as some traditional ethnic costumes, the performance degrades more significantly. This confirms that improving the reconstruction of fine-grained, high-frequency details---be they from complex patterns or unique materials---is a critical challenge. In our future work, we will explore injecting fine-grained clothing features directly into the attention layers of the diffusion model to supplement this high-precision information. Finally, we must acknowledge the computational cost associated with our diffusion-based approach. The training of our NCT framework takes approximately 80 hours on a single NVIDIA RTX 3090 GPU. For inference, generating a single image at 512x512 resolution with 50 DDIM steps requires about 15 seconds. While this performance is on par with other high-quality, controllable diffusion models, the inference latency is a significant limitation for real-time applications. We consider the exploration of model acceleration techniques, such as model distillation or quantization, to be a critical next step to enhance the practicality and user experience of the NCT framework.

\section{Conclusion}
The NCT is a novel method proposed to address the new challenges in the Cu-VTON task. This method aims to incorporate semantic descriptors of the target clothing as supplementary information and establish a learning framework to align these descriptions with visual images. These aligned features are seamlessly integrated into the diffusion model as conditional inputs. To ensure the preservation of clothing details and effective pose manipulation, an SC module is introduced to skillfully control the generation process. Experimental results vividly demonstrate the superiority of the proposed NCT model compared to existing methods in the Cu-VTON task. This highlights the effectiveness of the proposed approach in achieving substantial performance improvements in this field.

\section*{Credit Author Statement}
\textbf{Zhijing Yang:} Design the idea, help implement the algorithm, write and edit the manuscript. 
\textbf{Weiwei Zhang:} Implement the algorithm, conduct experiments, help write and edit the manuscript.
\textbf{Mingliang Yang:} Discuss the idea and help conduct experiments.
\textbf{Siyuan Peng:} Discuss the idea and implementation, help revise and edit the manuscript.
\textbf{Yukai Shi:} Discuss the idea and implementation, help revise and edit the manuscript.
\textbf{Junpeng Tan:} Help conduct new experiments and assist in revising and editing the manuscript.
\textbf{Tianshui Chen:} Design the idea and experiment validation settings, help write and edit the manuscript.
\textbf{Liruo Zhong:} Help revising and editing the manuscript, provide resource and supervision.

\section*{Data availability}
Data will be made available on request.

\section*{Declaration of competing interest}
The authors declare that they have no known competing financial interests or personal relationships that could have appeared to influence the work reported in this paper.

\section*{Acknowledgements}
This work was supported in part by the Natural Science Foundation of Guangdong Province under Grant No. 2025A1515010454, and in part by the National Natural Science Foundation of China (NSFC) under Grant No. 62206060.


\bibliographystyle{cas-model2-names}

\bibliography{paper-v2.bib}



\end{document}